\documentclass{egpubl}
\usepackage{pg2020}
 
\SpecialIssuePaper         

\usepackage[T1]{fontenc}
\usepackage{dfadobe}  

\usepackage{cite}  
\BibtexOrBiblatex
\electronicVersion
\PrintedOrElectronic
\ifpdf \usepackage[pdftex]{graphicx} \pdfcompresslevel=9
\else \usepackage[dvips]{graphicx} \fi

\usepackage{egweblnk} 

\usepackage{amsmath}
\usepackage{bm}
\usepackage{color}
\usepackage{threeparttable}
\usepackage{hyperref}
\usepackage[english]{babel}

\def\etal{\emph{et al}.~}
\def\eg{\emph{e.g}.~}
\def\ie{\emph{i.e}.~}

\begin{document}

\title[Human Pose Transfer by Adaptive Hierarchical Deformation]%
{Human Pose Transfer by Adaptive Hierarchical Deformation}


\author[Jinsong Zhang \emph{et al.}]
{\parbox{\textwidth}{\centering 
Jinsong Zhang\footnotemark[1],
Xingzi Liu\footnotemark[1] 
and Kun Li \footnotemark[2]
} 
\\
{\parbox{\textwidth}{\centering Tianjin University, Tianjin 300350, China.\\
}
}
}


\teaser{
\includegraphics[width=1.0\linewidth]{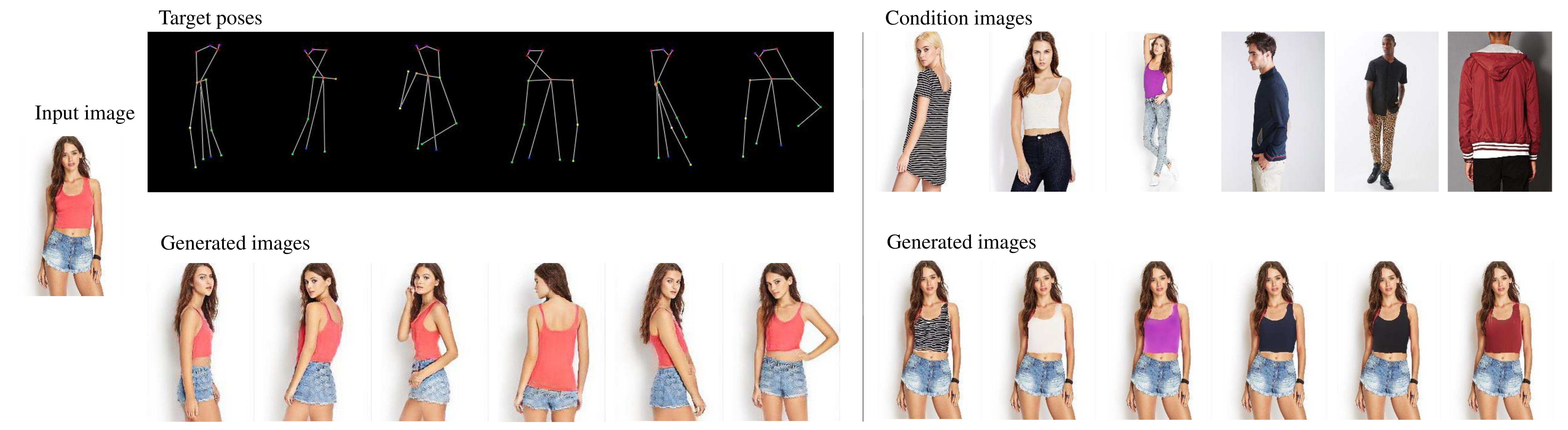}
\centering
\caption{Our method can generate person images in different target poses (left) and transfer upper clothing textures to a person image (right).}
\label{fig1}}

\maketitle
\begin{abstract}
Human pose transfer, as a misaligned image generation task, is very challenging. Existing methods cannot effectively utilize the input information, 
which often fail to preserve the style and shape of hair and clothes. In this paper, we propose an adaptive human pose transfer network with two hierarchical 
deformation levels. The first level generates human semantic parsing aligned with the target pose, and the second level generates the final textured person image 
in the target pose with the semantic guidance. To avoid the drawback of vanilla convolution that treats all the pixels as valid information, we use gated 
convolution in both two levels to dynamically select the important features and adaptively deform the image layer by layer. 
Our model has very few parameters and is fast to converge. 
Experimental results demonstrate that our model achieves better performance with more consistent hair, 
face and clothes with fewer parameters than state-of-the-art methods. Furthermore, our method can be applied to clothing texture transfer. The code is available for research purposes at \href{https://github.com/Zhangjinso/PINet_PG}{https://github.com/Zhangjinso/PINet\_PG}.


\end{abstract}  

\footnotetext[1]{Contribute equally to this work.}
\footnotetext[2]{Corresponding author: lik@tju.edu.cn }
\section{Introduction}

Human pose transfer aims to synthesize a person image from a source pose to a target pose while preserving the appearance details, 
which has potential applications in video generation~\cite{Walker-2017-103533}, person re-identification~\cite{qian2018pose, zheng2019joint} and image-based animation~\cite{Li2017SPA}. 
It is a very challenging and ill-posed problem due to misaligned transformation, occlusion, and high variance in poses.

Existing methods can be categorized into direct deformation and flow/transformation-based methods. 
Direct deformation method \cite{NIPS2017_6644} concatenated the source image with its condition and target pose as the inputs of the generator to synthesize the target image 
by adversarial learning. However, without considering spatial correspondences, the results of this method are a little blurry. 
Zhu \etal \cite{Zhu_2019_CVPR} proposed a progressive pose transfer network which utilized pose features to guide the image feature transfer, 
but the results often lost details (\emph{e.g.} hair style) due to using vanilla convolution in the encoder and decoder.  
More methods estimated flow or transformation matrix to guide the image generation. 
Siarohin \etal \cite{Siarohin_2018_CVPR} computed an affine transformation matrix based on the condition pose and the target pose to transform the image features. 
Dong \etal \cite{softgated} designed a soft-gated warping-block to learn feature-level mapping with the guidance of segmentation map, but the results rely on
the accuracy of estimated transformation matrix. Moreover, the matrix uses warped image features and is implemented only once, 
which is difficult to generate reasonable results for unknown regions.  
Han \etal \cite{clothflow} proposed a three-stage framework by adding a rendering network at the final stage to avoid the artifacts induced by wrong flows.
However, these flow/transformation-based methods are difficult to deal with large transformation between source image and target image.

All the above methods use vanilla convolution that treats all pixels as valid information, which cannot select significant regions to deform 
and usually have a large number of parameters. 
Vanilla convolution benefits aligned generation task by extracting alignment information, but for unaligned generation task, \eg, human pose transfer, 
the ability to select key areas to deform is more important.
Moreover, human pose transfer, as a challenging image deformation problem, is difficult to successfully generate the target image with only one warping, 
especially for large deformation. It is more reasonable to adaptively select important information and gradually deform the image from coarse to fine.



In this paper, we propose a hierarchical deformation framework for learning-based human pose transfer. Unlike flow/transformation-based methods that 
generate the image in target pose with only one deformation, we use an encoder-decoder architecture with gated convolutions to dynamically select important features 
and adaptively deform the image layer by layer. To simplify the challenging misaligned problem, we design a hierarchical deformation framework to transfer 
the human pose from coarse to fine, which includes a parsing generator and an image generator. 
The parsing generator generates human semantic parsing aligned with the target pose, and the image generator 
generates the final textured person image in the target pose with the semantic guidance to retain the clothing style and texture.
We conduct ablation study to verify our hypothesis. Comparative results with state-of-the-art methods demonstrate that our method achieves 
better human pose transfer results with fewer parameters. Our method can also be applied to image editing tasks, \eg clothing texture transfer.
Figure \ref{fig1} shows some examples generated by our method.

Our main contributions are summarized as follows:
\begin{itemize}

\item We propose a human pose transfer network with two hierarchical deformation levels. The first level generates human semantic parsing aligned with the target pose, 
and the second level generates the final textured person image in the target pose by fine deformation. This provides a coarse-to-fine deformation framework and 
alleviates the difficulty of direct deformation from source to target.

\item We introduce gated convolution to avoid the drawback of vanilla convolution that treats all the pixels as valid information. 
This copes well with the unaligned image generation task by learning a dynamic feature selection mechanism and adaptively deforming the image layer by layer. 

\item Our model has very few parameters and is fast to converge.

\item Our model can be applied to image editing tasks, \eg clothing texture transfer. 

\end{itemize}

The rest of this paper is organized as follows. 
Section  \ref{sec:related} presents a brief review of related work. 
Section \ref{sec:method} describes the proposed network. 
Experimental results are presented in Section \ref{sec:experiment}, and the paper is concluded in Section \ref{sec:conclude}.

\section{Related Work}  
\label{sec:related}

\subsection{Person Image Generation}

Lassner \emph{et al.} \cite{Lassner:GeneratingPeople:2017} combined variational auto-encoder \cite{kingma2013auto-encoding} and Generative Adversarial Network (GAN) to 
generate random person images with different appearance for full body. Zhu \emph{et al.} \cite{Zhu2018CVPR}  proposed a novel pipeline for synthesizing human bodies 
from monocular image. Balakrishnan \emph{et al.} \cite{Balakrishnan2018CVPR}  decomposed human image generation tasks into multiple foregrounds with different body parts 
and backgrounds. Si \emph{et al.} \cite{Si_2018_CVPR} proposed a pose converter network, in which the foreground converter network and the background converter network 
use a multi-stage confrontation loss to generate more realistic images. Several previous research work \cite{Han_2018_CVPR,lahner2018deepwrinkles,wang2018toward} 
focused on virtual try-on applications and made great progress in transferring clothes for a given character image, but the pose and shape of the character was unchanged. 
Unlike these methods which generate person images with unchanged pose, we propose a new person image generation method with various poses and viewpoints, 
which can also change the texture of clothing.

\begin{figure*}[!ht]
\centering
\includegraphics[width=0.9\linewidth]{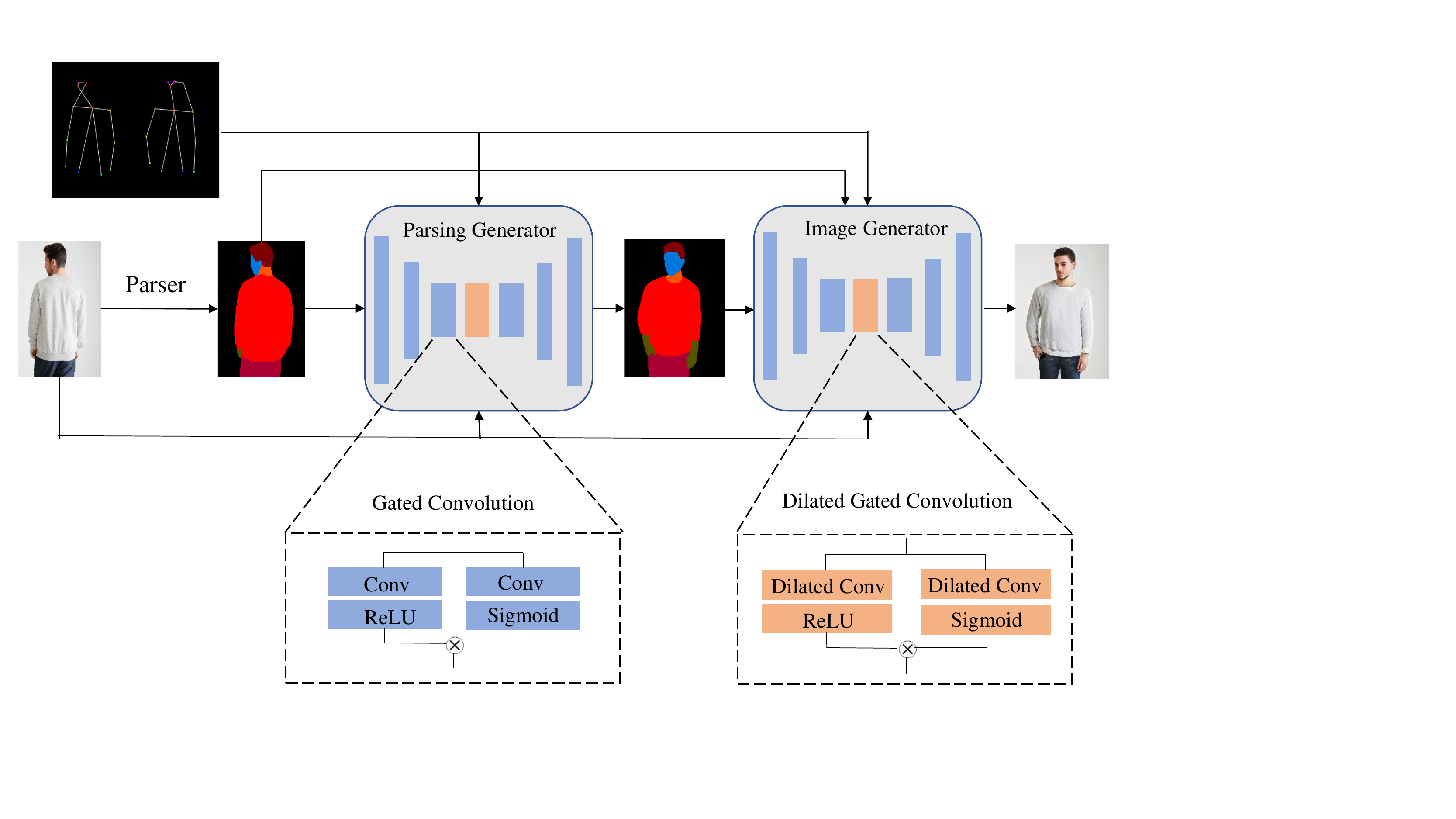}
\caption{Overview of our framework. Given an input person image together with the source and target poses, our method generates the person image in the target pose by 
hierarchical deformation with a parsing generator and an image generator. We use gated convolution to dynamically select important features and adaptively deform the image layer by layer.}
\label{network}
\end{figure*}

\subsection{Human Pose Transfer}

Human pose transfer can synthesize a new image with changed poses from a single person image. 
Several methods used 2D keypoints as pose representation to directly synthesize the target image.
Ma \emph {et al.} \cite {NIPS2017_6644} proposed the first human pose transfer framework,
which generated a coarse result and then refined it. However, this model is computationally inefficient and complicated to train. 
Ma \emph{et al.} \cite{Ma_2018_CVPR} improved their previous work by using a decomposition strategy. 
Esser \emph{et al.}~\cite{Esser_2018_CVPR} adopted variational autoencoder to sample appearance, 
and used U-Net \cite{ronneberger2015u} to keep shape information, for interactive modeling. 
Li \emph{et al.} \cite{Li_2019_CVPR} introduced 3D flow graph with conditional poses, target poses 
and the visibility map to guide the transformation of image features and pixels. 
Zhu \emph{et al.} \cite{Zhu_2019_CVPR} proposed to deform image features 
in the latent space progressively. 
However, these method used 2D keypoints as pose representation, which is difficult to extract semantic correspondence directly between the source and target images.
Some methods used human parsing maps as semantic guidance and estimated a transformation matrix or flow to synthesize the target image. 
Dong \emph{et al.} \cite{softgated} first generated human parsing map and predicted a transformation matrix with the help of human parsing maps aligned with 
the source pose and the target pose, and then deformed image features using the transformation matrix. 
Song \emph{et al.} \cite{song2019unsupervised} proposed an unsupervised method to synthesize human parsing map and the target image by designing a cycle loss. 
Han \emph{et al.} \cite{clothflow} used synthesized parsing map to estimate a cloth flow mapping, warped image features to generate the clothing image without body, 
and concatenated the source image to synthesize the final result. 
Dong \emph{et al.} \cite{dong2019towards} proposed a similar method with coarse-to-fine stage to enhance image details. 
Hsieh \emph{et al.} \cite{hsieh2019fashionon} decomposed this task into three stages: pose-guided parsing translation, segmentation region coloring, and salient region refinement. 
Unlike these methods that use vanilla convolution, in this paper, we adopt gated convolution to learn a dynamic feature selection mechanism and adaptively deform the image layer by layer. 
Instead of estimating a transformation matrix or flow to deform features, 
we propose to extract and deform image features at the same time, which can preserve more important information and predict unknown regions.

\subsection{Feature-wise Gating}
Feature-wise gating is widely used in speech \cite{oord2016wavenet}, language \cite{dauphin2017language}, and vision \cite{hu2018squeeze, van2016conditional, wang2017gated}. WaveNets \cite{oord2016wavenet} achieved promising results by applying a special feature gating $O_{x,y} = \tanh(\emph{w}_1\emph{x}) \odot softmax($\emph{w}$_{2}$\emph{x}$)$ to model audio signals. Gated pixelCNN \cite{van2016conditional} also used this feature gating to model vision signals. Yu \emph{et al.}\cite{yu2019free} introduced this formulation as gated convolution into the image inpainting task, which significantly improves the inpainting quality of free-form masks and inputs. Inspired by \cite{yu2019free}, we adopt gated convolution to deal with the unaligned image generation task.

\section{Our Approach}
\label{sec:method}
As shown in Figure \ref{network}, we propose a semantic-guided human pose transfer network with two hierarchical deformation levels. 
In the first level, parsing generator $G_p$ generates a human parsing map aligned with desired pose. 
In the second level, image generator $G_i$ synthesizes the final textured person image in the target pose with the semantic guidance. 
Details of our network architecture can be found in the supplementary video.

We apply Human Pose Estimator (HPE) \cite{cao2017realtime} to extract 18 heatmaps as pose representation. 
An off-the-shelf human parser \cite{gong2018instance} is used to produce human segmentation maps with 20 labels. 
Because the predicted segmentation maps by human parser have some ambiguous parts (\emph{e.g.}, left leg and right leg), 
we re-organize the map into 12 categories: background, hair, upper clothes, dress, pants, neck, skirt, face, hands, legs, shoes and hat.    

In this section, we first explain the motivation to use gated convolution in both parsing generator and image generator, and then describe the details of our model. 
Finally, we show an application of texture transfer with our model.

\subsection{Gated Convolution}

Vanilla convolutions are widely used in convolution neural networks, which achieve great progress in object detection, image segmentation, and image-to-image translation. 
All the existing methods for human pose transfer used vanilla convolutions to comprise their models. The vanilla convolution is formulated as
\begin{equation}
O_{y,x} = \sum_{i=-k_h}^{k_h}\sum_{i=-k_w}^{k_w} W_{k_h+i,k_w+j} \cdot I_{y+i,x+j},
\end{equation}
where $k_h$ = $\frac{k_{sh}-1}{2}$ and $k_w$ = $\frac{k_{sw}-1}{2}$ in which $k_{sh}$ and $k_{sw}$ are the kernel sizes (\emph{e.g.}, 3 $\times$ 3). 
$W$ represents convolutional filters.

The formulation of vanilla convolution layers indicates that the output values in all spatial location are calculated with the same filter. 
It takes all pixels as valid values and extracts local features with a sliding window, which makes sense to object detection, image segmentation and aligned generation tasks. 
However, for misaligned tasks, \emph{e.g.}, human pose transfer, the features extracted by vanilla convolution do not always have a positive impact on the output.
Therefore, it is more important to learn a dynamic feature selection mechanism to deform the image.
Inspired by \cite{yu2019free}, we introduce gated convolution into human pose transfer, which is formulated as

\begin{equation}
\begin{split}
O_{x,y} = \phi(\sum_{i=-k_h}^{k_h}\sum_{i=-k_w}^{k_w} u_{k_h+i,k_w+j} \cdot I_{y+i,x+j}) \odot \\ \sigma(\sum_{i=-k_h}^{k_h}\sum_{i=-k_w}^{k_w} v_{k_h+i,k_w+j} \cdot I_{y+i,x+j}),
\end{split}
\end{equation}
where $\sigma$ denotes sigmoid function and the output gating values are between 0 and 1. 
$\phi$ denotes any activation function (\emph{e.g.}, \emph{Tanh} in WaveNet \cite{oord2016wavenet}). $u$ and $v$ are two different convolutional filters. 
We use LeakyReLU \cite{maas2013rectifier} as $\phi$ in our model.

For the parsing generator, gated convolution can obtain useful information at each spatial location, which is suitable for preserving the semantic parts of the person 
(\emph{e.g.}, clothing style). For the image generator, gated convolution can preserve important features and deform key areas to generate textured person image. 
Therefore, we replace all vanilla convolutions with gated convolutions to adaptively extract and deform features.

\subsection{Parsing Generator}
For human pose transfer, the source image and the target image contain the same person with the same clothes and body shape, 
and hence we design a human parsing generator to build the semantic correspondence between them.  
Different from previous work, we use gated convolution to compose our parsing generator.

Because the misalignment between the input image and the target image, we design an encoder-decoder architecture for the parsing generator. Table \ref{tab:tab_net} shows the details of our network architecture, where $c_{out}$ is the dimension of the output (12 for parsing generator and 3 for image generator). 
Given an input person image $I_s$ and a target pose $P_t$, the parsing generator learns to generate the human parsing map $M_g$ conditioned on image $I_s$ and pose $P_t$. 
Specifically, we first extract source pose $P_s$ and source parsing map $M_s$ from the input person image using a human pose estimator \cite{cao2017realtime} 
and a human parser \cite{gong2018instance}, respectively. Then, we concatenate them with source image $I_s$ as the input of our parsing generator. 
The processing of our parsing generator $G_p$ can be written as 
\begin{equation}
M_g = G_p(I_s, P_s, P_t, M_s).
\end{equation}
The reconstruction loss of $G_p$ is defined as $\ell_1$ distance loss between target parsing map $M_t$ and generated parsing map $M_g$:
\begin{equation}
\mathcal{L}_{\ell_1}=||M_g-M_t||_1.
\end{equation}
We also apply the categorical cross-entropy loss to encourage the generator to synthesize high-quality parsing maps:	
\begin{equation}
\mathcal{L}_{\log} = \mathcal{L}_{\log}(M_t, M_g) =  - \frac{1}{N} \sum_{i=0}^{N-1}  M_{t_i}  \log (S(M_{g_i}),
\end{equation}
where $N$ is the number of categories of labels ($N=12$ in our case), and $S$ denotes softmax function.
The final loss function of our parsing generator can be formulated as:
\begin{equation}
\mathcal{L}_{parsing} = \mathcal{L}_{\log} + \mathcal{L}_{\ell_1}.
\end{equation}

\begin{table}[!t]
\renewcommand{\arraystretch}{1.0}
\small
\setlength{\tabcolsep}{1.2mm}
\begin{center}

\caption{Details of our network architecture.}\label{tab:tab_net}
\begin{tabular}{ccccc}
\hline
Operation & Kernel Size & Stride & Dilation & Output Shape \\
GatedConv & 7x7 & 1 & 1& (256, 176, 64)\\
GatedConv & 3x3 & 2 & 1& (128, 88, 64)\\
GatedConv & 3x3 & 1 & 1& (128, 88, 128)\\
GatedConv & 3x3 & 2 & 1& (64, 44, 128) \\
GatedConv & 3x3 & 1 & 1& (64, 44, 256) \\
GatedConv & 3x3 & 1 & 1& (64, 44, 256) \\
GatedConv & 3x3 & 1 & 1& (64, 44, 256) \\
DilatedGatedConv & 3x3 & 1 & 2& (64, 44, 256) \\
DilatedGatedConv & 3x3 & 1 & 4& (64, 44, 256) \\
Self-attention Module & - & - & -& (64, 44, 256) \\
GatedConv & 3x3 & 1 &  1&(64, 44, 256) \\
GatedConv & 3x3 & 1 &  1&(64, 44, 256) \\
Upsample & - & - &  1&(128, 88, 256) \\
GatedConv & 3x3 & 1 &  1&(128, 88, 128)\\
GatedConv & 3x3 & 1 &  1&(128, 88, 128)\\
Upsample & - & - &  -&(256, 176, 128)\\
GatedConv & 3x3 & 1 & 1&  (256, 176, 64)\\
GatedConv & 3x3 & 1 & 1&  (256, 176, 32)\\
GatedConv & 7x7 & 1 & 1&  (256, 176,  $c_{out}$)\\

\hline
\end{tabular}	
\end{center}
\end{table}

\begin{figure*}[!htp]
\centering
\includegraphics[width=1.0\linewidth]{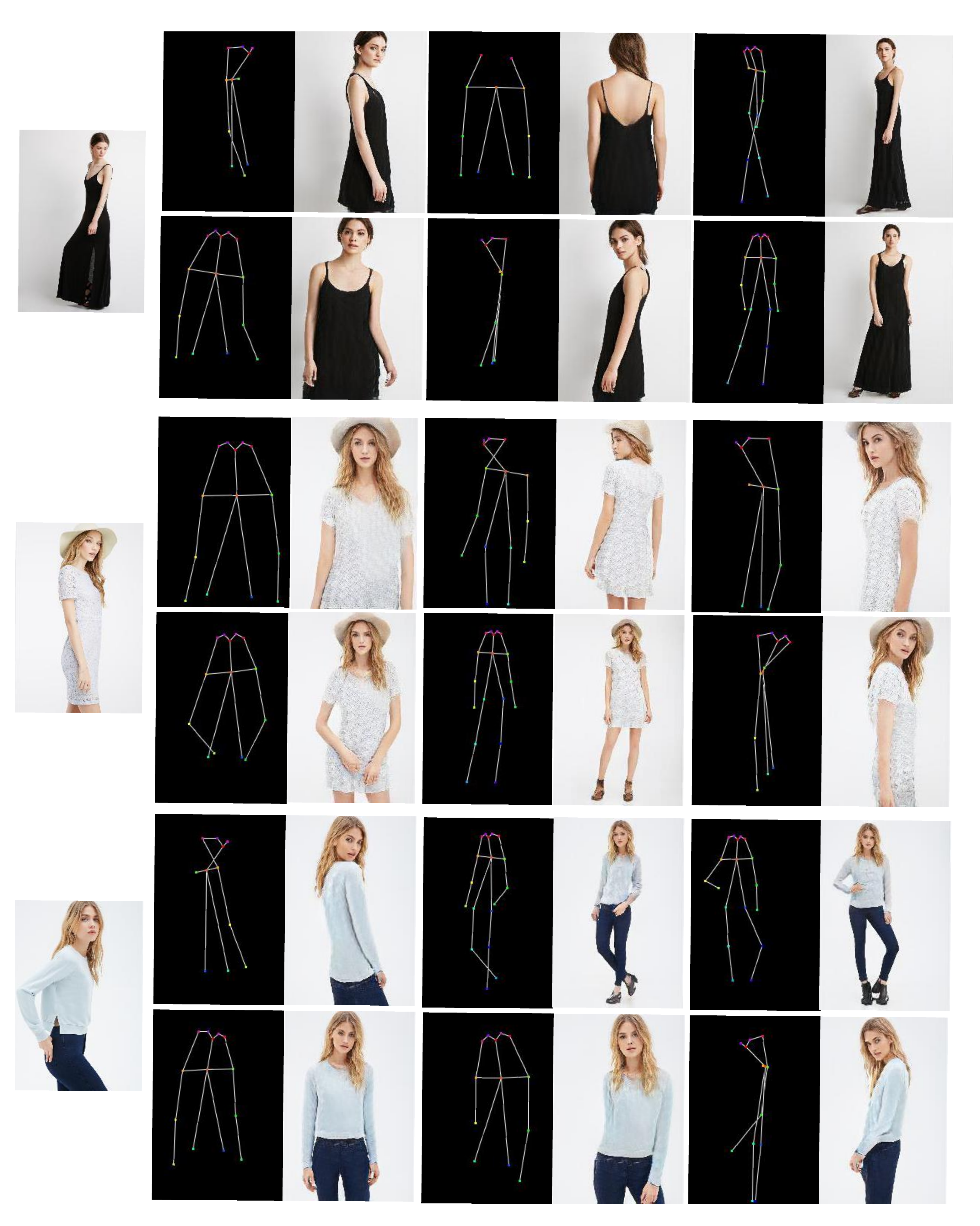}
\caption{Results of person image synthesis in different poses using our method.}
\label{arbb}
\end{figure*}

\begin{figure*}[!ht]
\centering
\includegraphics[width=0.95 \linewidth]{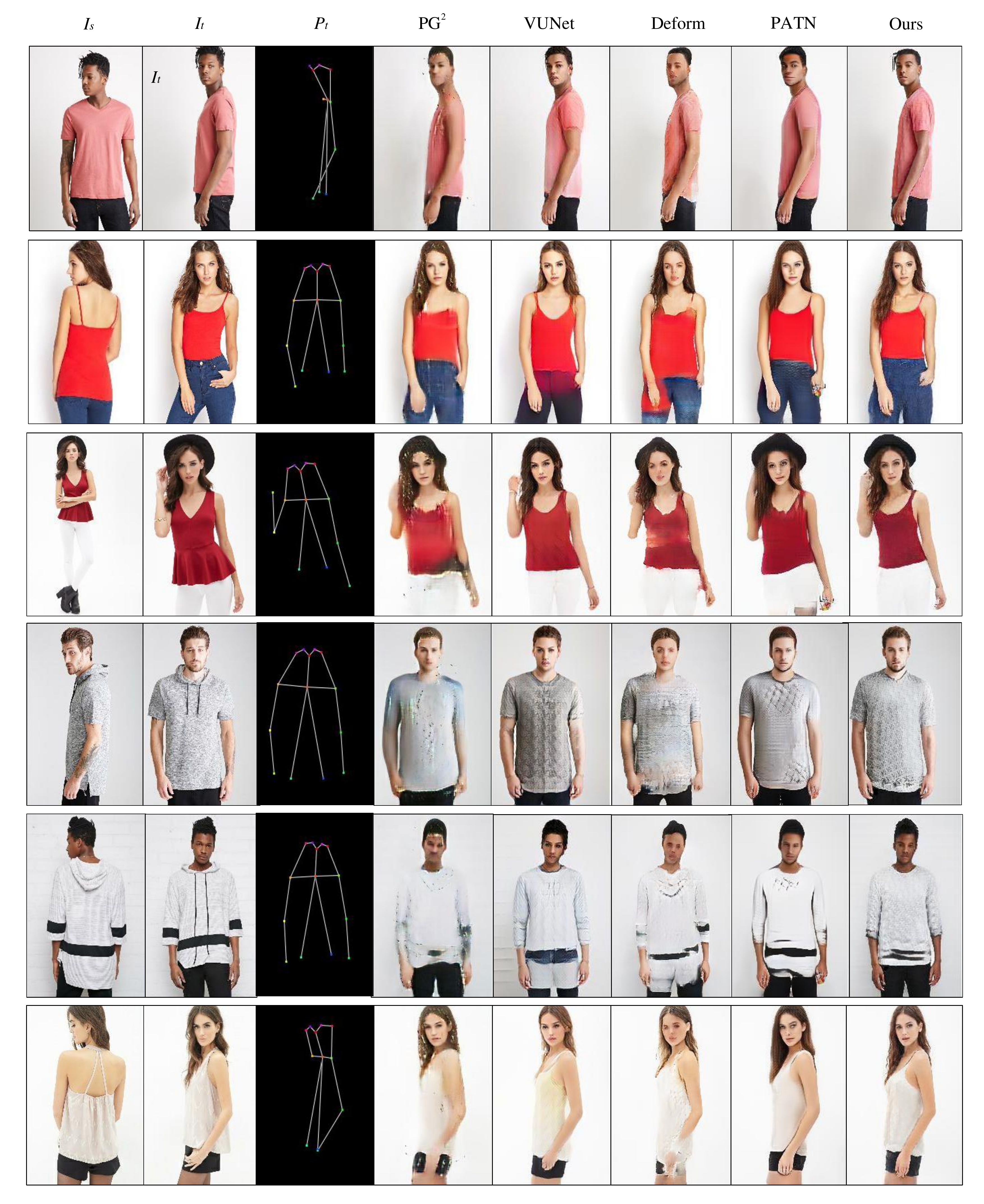}
\caption{Qualitative results on DeepFashion dataset compared with PG$^2$\cite{NIPS2017_6644}, VUnet\cite{Esser_2018_CVPR}, Deform\cite{Siarohin_2018_CVPR} 
and PATN\cite{Zhu_2019_CVPR}.}
\label{com1}
\end{figure*}

\subsection{Image Generator}
With the source human parsing map $M_s$ and the generated human parsing map $M_g$, previous work \cite{softgated, clothflow} used human parsing maps 
to estimate a flow mapping and warped image features to generate the results using this flow mapping. 
However, their models may lose clothing style due to imprecise flow mapping and generate artifacts in some unknown regions. 
Instead of estimating a flow mapping using human parsing maps, we use gated convolution to extract image feature and semantic correspondence, and deform the image feature. 
The image generator aims to deform the source image $I_s$ based on the parsing maps $M_s$ and $M_g$ by extracting the semantic correspondence. 
We also feed the target pose $P_t$ into our image generator to encourage the synthesized image to be aligned with the target pose when the generated parsing map is not precise.

We adopt the same encoder-decoder architecture for the image generator. 
Based on a pair of parsing maps and the source person image, the encoder is responsible for extracting semantic correspondences and encoding the image, 
while the decoder is used to refine and decode the final feature to deliver the result. The dilated gated convolution \cite{YuKoltun2016} is applied to expand the receptive filed. 

The generative adversarial framework \cite{NIPS2014_5423} is used to generate more realistic images by mimicking the distributions of the ground truth $I_t$. 
In traditional GAN, the discriminator is used to judge whether the generated image is real or fake to encourage the generator to synthesize realistic images. For conditional image generation task, it is also important to make the generated images meet the requirement of condition image (e.g., the pose of the generated image should be aligned with the target pose in human pose transfer), in addition to generating realistic images. Therefore, we use two discriminators: pose discriminator $D_P$ and appearance discriminator $D_A$, to encourage the generator to synthesize images aligned with the target pose 
and preserve the texture consistent with the input image. The conditional adversarial loss is defined as: 

\begin{equation}
\begin{split}
\mathcal{L}_{CGAN} &=E\{\log[D_A(I_s,I_t)\cdot D_P(I_t,P_t)]\} + \\
& E\{\log[(1-D_A(I_s,I_g))\cdot (1-D_P(I_g,P_t))]\}.
\end{split}	
\end{equation}
We use $\ell_1$ distance loss between the generated image $I_g$ and the ground truth $I_t$, which is defined as:	
\begin{equation}
\mathcal{L}_{\ell_1}=||I_g-I_t||_1.
\end{equation}	
Perceptual loss \cite{johnson2016perceptual} has achieved great success in image synthesis\cite{isola2017image,zhu2017unpaired,liu2019few}. 
We apply a perceptual loss $\mathcal{L}_{percep}$ to compute the distances of high-level features in the pre-trained model between the generated image $I_g$ and the ground truth $I_t$. 
We formulate the perceptual loss as:
\begin{equation}
\mathcal{L}_{percep}=\sum_{i=1}^{N}\alpha_i||\phi_i(I_g)-\phi_i(I_t)||_1,
\end{equation}
where $\phi_i(I_g)$ denotes the feature map of the $i$-th ($i=0, 1, 2, 3, 4$) layer in the pre-trained network $\phi$ for the generated image $I_g$.
We use VGG-19 model \cite{simonyan2014very} pre-trained on ImageNet \cite{deng2009imagenet} as $\phi$ and extract feature maps from layers $relu$\{1\_1, 2\_1, 3\_1, 4\_1, 5\_1\}.

The final loss function is defined as :
\begin{equation}
\mathcal{L}_{Image} = \lambda_1 \mathcal{L}_{\ell_1} + \lambda_2 \mathcal{L}_{percep} + \lambda_3 \mathcal{L}_{CGAN},
\end{equation}
where $\lambda_1$, $\lambda_2$, $\lambda_3$ represent the weights of $ \mathcal{L}_{\ell_1}$, $\mathcal{L}_{percep}$, $\mathcal{L}_{CGAN}$ that contribute to $\mathcal{L}_{Image}$, respectively.

\subsection{Implementation Details}

For two generators, we first train the parsing generator and the image generator for about 60K iterations, respectively. 
The input of parsing generator is a triplet ($I_s$ ,$P_s$, $P_t$, $M_s$), and the output is a generated parsing map aligned with target image. 
For the image generator, we alternatively train the generator and the discriminators. The image generator takes a triplet ($I_s$, $M_s$, $M_t$, $P_t$) as input 
and delivers the generated image $I_g$. To train the discriminators, the appearance discriminator $D_A$ takes ($I_s$, $I_t$) and ($I_s$, $I_g$) as inputs, 
and the pose discriminator $D_p$ takes ($P_t$, $I_t$) and ($P_t$, $I_g$) as inputs. 
Then, we train two generators jointly for around 90K iterations. The initial learning rate is linearly decayed to 0 after 50K iterations.

The coefficients in the loss function of image generator ($\lambda_1$, $\lambda_2$, $\lambda_3$) are set to (1, 0.5, 5). 
Spectral Normalization \cite{Yoshida2018Spectral} is adopted after every convolution layer in two discriminators to improve the stability of training process. 
Adam optimizer~\cite{kingma2014adam} with $\beta_1$=0.5 and $\beta_2$=0.999 is applied to train our model.

\subsection{Texture Transfer}
\label{sec_trans}
Because we use human parsing map as an intermediate result to represent semantic correspondences, 
we can achieve texture transfer by replacing the source body parts with new clothing texture, utilizing human parsing maps. 
For example, we can replace the texture of upper clothes in the source image $I_s$ with that in the condition image $I_c$. To achieve this, 
we first take the condition image $I_c$, parsing maps $M_c$ and $M_s$ extracted from $I_c$ and $I_s$ and the source pose $P_s$ as inputs to our image generator 
to synthesize a new image with the original body shape of the source image and the texture of the condition image. 
Then, we crop the generated image by the body part mask $M_{s_i}$ ($i \in [1, 12]$) from human parsing map $M_s$. 
Finally, we deliver the result by replacing the region of  $M_{s_i}$ in the source image $I_s$ with the generated image we cropped. 
Formally, the texture transfer process is formulated as
\begin{equation}
I_f = I_s \odot (1- M_{s_i}) + G_I(I_c,M_c,M_s,P_s) \odot M_{s_i},
\end{equation} 
where $G_I$ is our image generator. More details can be found in the supplementary video. 

\begin{figure}[!t]
\centering
\includegraphics[width=0.94 \linewidth]{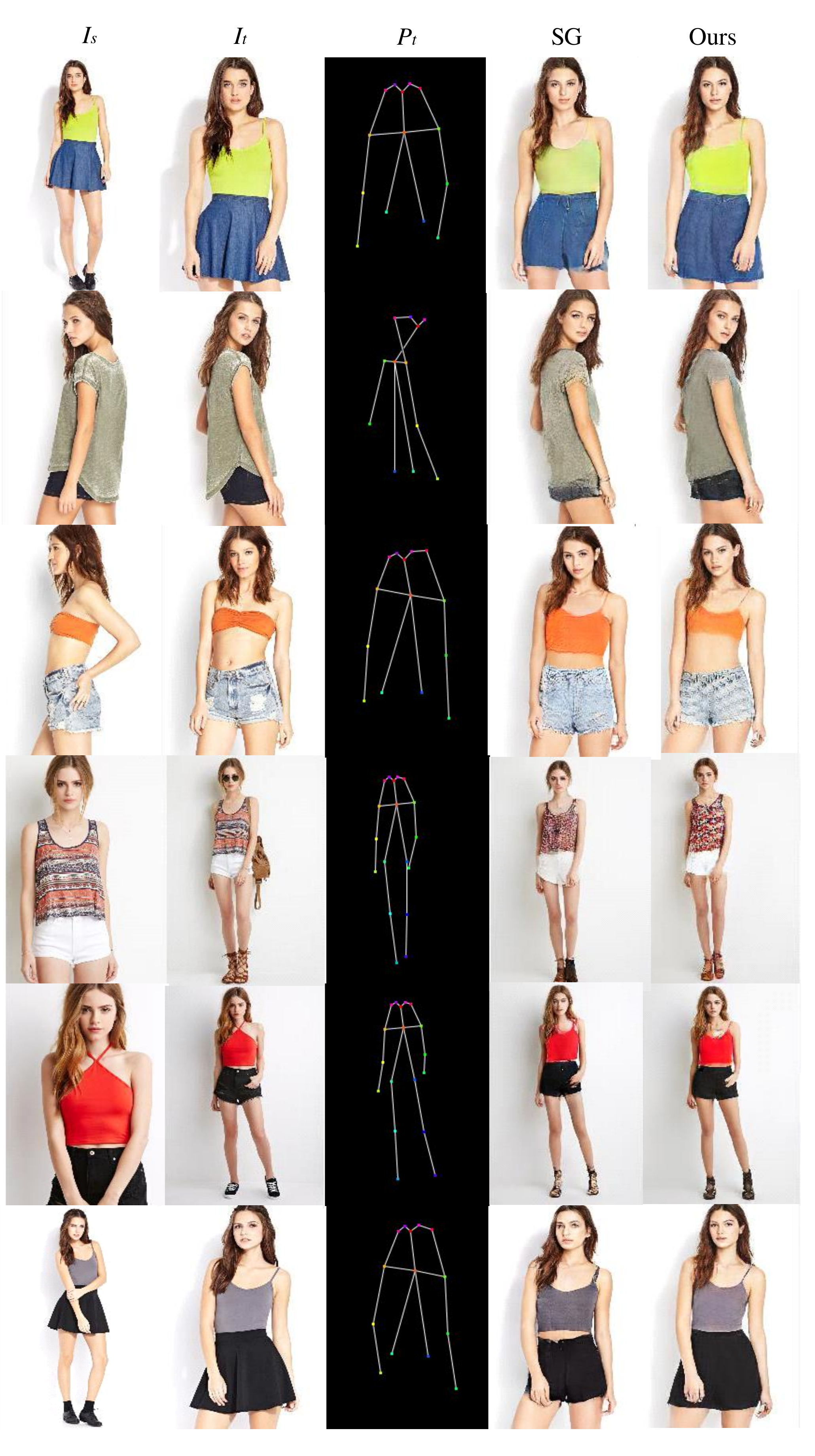}
\caption{Qualitative results on DeepFashion dataset compared with SG \cite{softgated}. Please zoom in for details.}
\label{comsoft}
\end{figure}

\begin{figure}[!t]
\centering
\includegraphics[width=1.0 \linewidth]{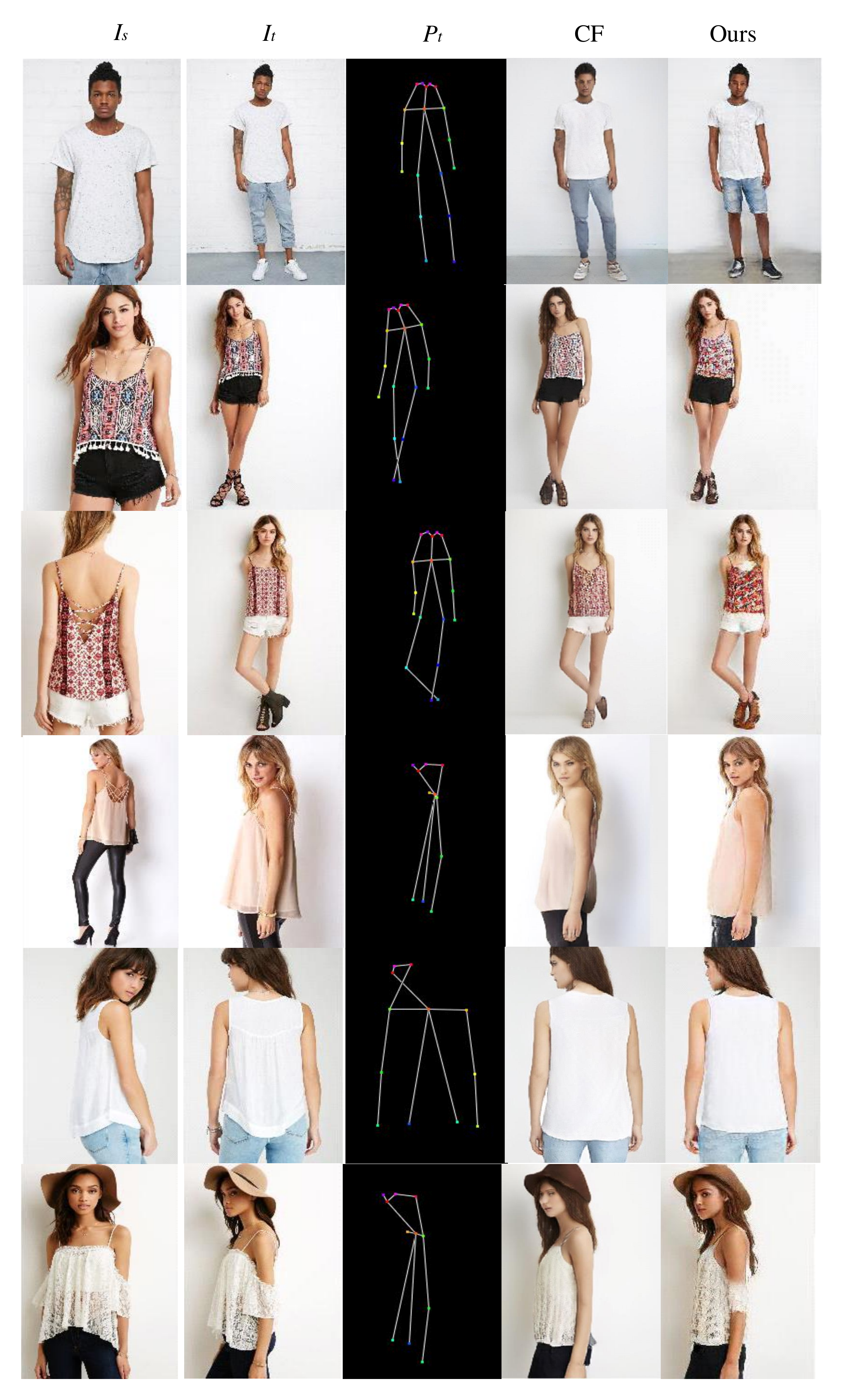}
\caption{Qualitative results on DeepFashion dataset compared with CF \cite{clothflow}. Please zoom in for details.}
\label{comcloth}
\end{figure}

\section{Experimental Results}
\label{sec:experiment}

In this section, we first give some human pose transfer results of our method, and quantitatively and qualitatively compare our method with several state-of-the-art methods. 
Then, we perform ablation study to verify the effect of different components in our model. Finally, we present an application in texture transfer. 
More results can be found in the supplementary video.

%


\subsection{Results}
To verify the effectiveness of our method, we generate person images with the same source person image and several different target poses on the 
DeepFashion dataset \cite{liu2016deepfashion} that has large variation in pose and appearance. 
This dataset contains 52712 images with the resolution of 256 $\times$ 256. Following PATN \cite{Zhu_2019_CVPR}, we adopt the same data configuration, 
collecting 101966 training pairs and 8570 testing pairs. Note that the person identities in the test set are different from those in the training set.
The source and target poses are estimated by a human pose estimator \cite{cao2017realtime}.  
Figure \ref{arbb} shows some results using our method. 
Our method can generate sharp details by preserving the textures (\emph{e.g.}, hair) and clothing patterns (\emph{e.g.}, dress and hat). 
Moreover, our results retain facial details. This appealing property attributes to our hierarchical deformation framework with gated convolutions.

\noindent
\subsection{Comparison}

\noindent
\textbf{Quantitative comparison.}  To evaluate the performance of human pose transfer, we use three metrics as our evaluation metrics. 
Inception Score (IS) is commonly used to measure the quality of image generation \cite{NIPS2017_6644}. 
To coincide  with  human  judgment, Learned Perceptual Image Patch Similarity (LPIPS) \cite{zhang2018perceptual} is used to 
calculate the reconstruction error between the generated image and the ground truth. 
Fr{$\acute{\textup{e}}$}chet Inception Distance (FID) \cite{heusel2017gans} is used to measure the realism of the generated images.
Table \ref{tab:tab_com} shows the quantitative results compared with four state-of-the-art methods: 
PG$^2$ \cite{NIPS2017_6644}, VUnet \cite{Esser_2018_CVPR}, Deform \cite{Siarohin_2018_CVPR} and PATN \cite{Zhu_2019_CVPR}. 
We run the pre-trained models of these methods and use the same data division as PATN \cite{Zhu_2019_CVPR}. 
For PG$^2$ \cite{NIPS2017_6644}, VUnet \cite{Esser_2018_CVPR} and Deform \cite{Siarohin_2018_CVPR}, 
our test images may appear in their training set due to lack of their data division details. 
As shown in Table \ref{tab:tab_com}, our method achieves the best performance in LPIPS and FID, which are more consistent with human judgement. Besides, our model has the fewest parameters.

\begin{table}[htbp]
\renewcommand{\arraystretch}{1.0}
\small
\setlength{\tabcolsep}{1.7mm}
\begin{center}
\caption{Quantitative comparison with four state-of-the-art methods on DeepFashion dataset.}\label{tab:tab_com}
\begin{tabular}{|c|c|c|c|c|c|}
\hline
{Model} & IS $\uparrow$ & LPIPS $\downarrow$ & FID $\downarrow$ & Parameters \\

\hline
PG$^2$ \cite{NIPS2017_6644}  & 3.163 &  0.2901 & 45.288 & 437.09M \\
VUnet \cite{Siarohin_2018_CVPR}  & 3.362 & 0.2637 & 23.667 & 139.36M\\
Deform \cite{Esser_2018_CVPR}  & \textbf{3.440} & 0.2330 & 18.457 & 82.08M\\
PATN \cite{Zhu_2019_CVPR}  & 3.209 & 0.2533 & 20.739 & 41.36M\\
\hline
Ours   & 3.419 &\textbf{0.2159} &  \textbf{12.635}  & 20.41M \\ 
\hline
\end{tabular}
\end{center}
\end{table}

\begin{figure*}[!ht]
\centering
\includegraphics[width=0.82\linewidth]{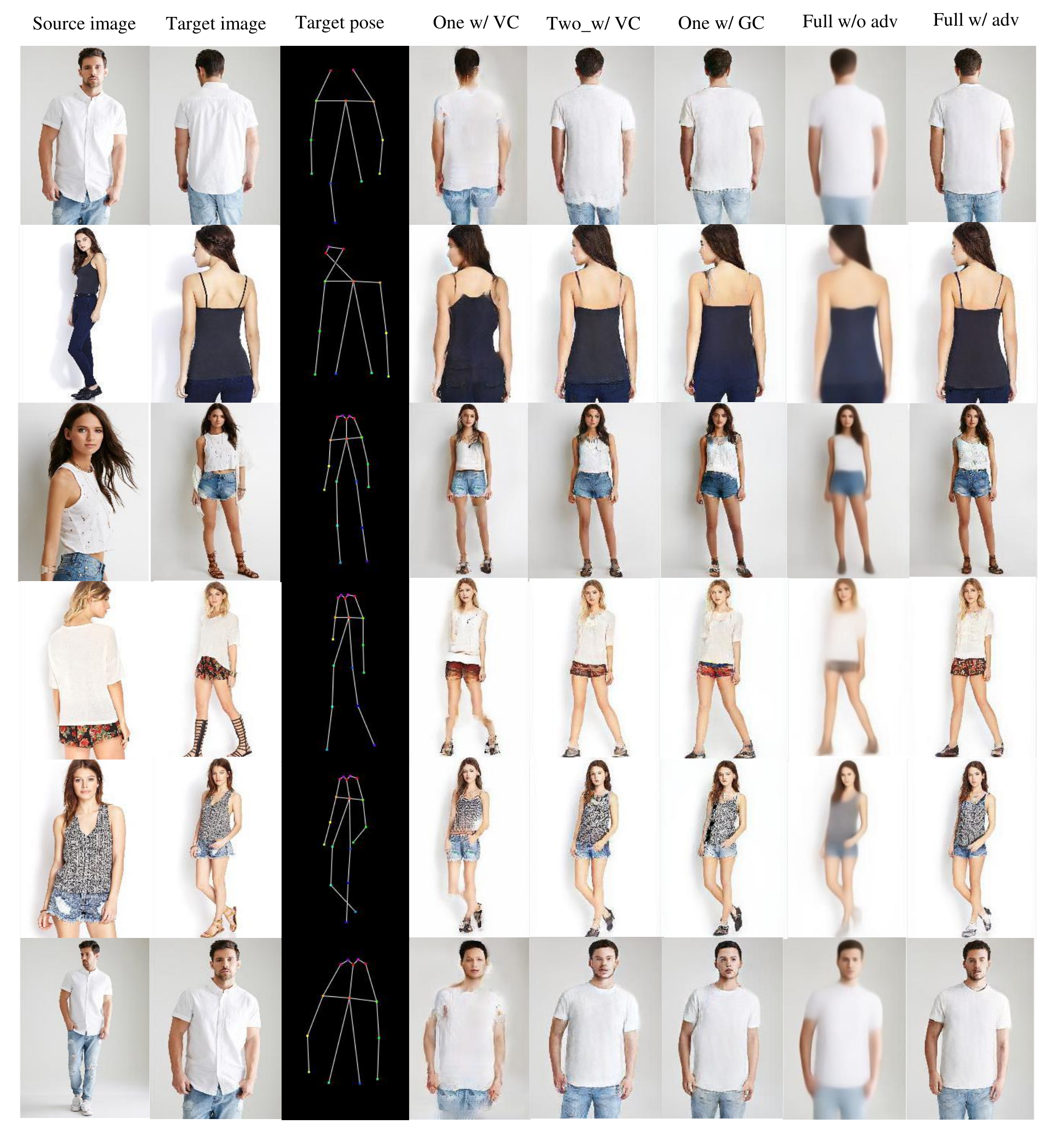}
\caption{Qualitative results of ablation study. The fourth and fifth columns indicate one-stage model and two-stage model with vanilla convolution respectively, and the sixth column indicates one-stage model with gated convolution. Last two columns denote the results of our full model with and without adversarial loss, respectively.}
\label{pic_aba}
\end{figure*}

\noindent
\textbf{Qualitative comparison.} Figure \ref{com1} shows qualitative results compared with PG$^2$ \cite{NIPS2017_6644}, VUnet \cite{Esser_2018_CVPR}, 
Deform \cite{Siarohin_2018_CVPR} and PATN \cite{Zhu_2019_CVPR}. Our model can generate more sharp images with rich details. For example, 
the hair in our generated images is more realistic and the style has been preserved. Furthermore, our model keeps shape consistence, \eg, 
the hat in the third row. The images synthesized by our model also have more consistent texture with source person images, \eg, 
the color of skin and clothes, which means that our image generator can transfer textures based on human parsing maps.

We also compare our method with two semantic-guided methods, SG \cite{softgated} and CF \cite{clothflow}, in Figure \ref{comsoft} and Figure \ref{comcloth}, respectively.
Due to lack of their codes, we use the results presented in their papers. 
As shown in Figure \ref{comsoft}, our model can preserve more clothing details and clothing style, \eg, dress in the first/last row and upper clothes in the third row. 
Besides, our model can generate more reasonable unknown regions with less artifacts, \eg, shoes in the fourth and fifth rows. 
As shown in Figure \ref{comcloth}, our model generates more consistent semantic shape and texture with the source image, \eg, hair in the first row and the third row.

\begin{figure*}[!htp]
\centering
\includegraphics[width=1.0\linewidth]{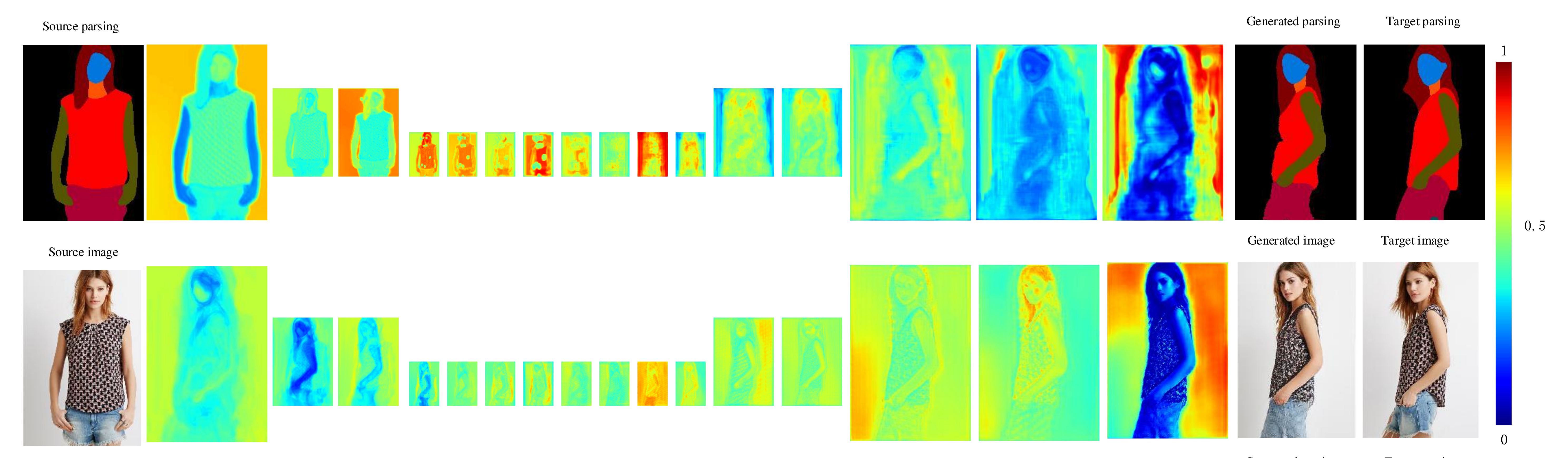}
\caption{Visualization of our attention mask in each gated convolution layer of the parsing generator and the image generator. }
\label{att}
\end{figure*}

\subsection{Ablation study}

We conduct ablation study to verify the effectiveness of the important parts. We train four ablation models compared with our model:

\noindent
\textbf{One w$\slash$ VC.} The one-stage model with vanilla convolution is similar with the first stage in PG$^2$ \cite{NIPS2017_6644}. However, we use two discriminators and losses as shown in Section \ref{sec:method}.

\noindent
\textbf{Two w$\slash$ VC.} The two-stage model with vanilla convolution is also a hierarchical deformation framework with a parsing generator and an image generator, 
while using vanilla convolution instead of gated convolution. Other settings are the same as our model.

\noindent
\textbf{One w$\slash$ GC.} The one-stage model directly synthesizes the results from the input image together with the source pose and the target pose without semantic guidance. 
We employ gated convolution in the generator and the same discriminators as illustrated in Section \ref{sec:method}.

\noindent
\textbf{Full w$\slash$o adv.} We retrain our model without adversarial loss to investigate the effect of adversarial loss.

Table \ref{tab_ab} and Figure \ref{pic_aba} give the quantitative and qualitative results compared with one stage model and vanilla convolution model.
As shown in Figure \ref{pic_aba}, compared with one-stage model based on gated convolution, our image generator generates more realistic images with details. 
This demonstrates that our image generator can extract richer semantic information and deform image features more reasonable with human parsing map as an intermediate result.
The vanilla convolution model also synthesizes reasonable results, but the texture and identity of the results are less consistent with the source images. 
On the contrast, our model with gated convolution can learn a dynamic selection mechanism and adaptively select important regions to deform. 
Moreover, compared with PG$^2$, one-stage model with vanilla convolution gets better results,
which illustrates the advantages of our two discriminators and reasonable losses. Besides, there is no doubt that adversarial loss encourages the model to synthesize realistic images.
The quantitative results given in Table \ref{tab_ab} also verify the rationality of our model.


\begin{table}[htbp]
\renewcommand{\arraystretch}{1.0}
\small
\setlength{\tabcolsep}{1.7mm}
\begin{center}
\caption{Quantitative comparison  of ablation study.}\label{tab_ab}
\begin{tabular}{|c|c|c|c|c|}
\hline
{Model}& IS $\uparrow$ & LPIPS $\downarrow$ & FID $\downarrow$  \\
\hline
One w$\slash$ VC & 3.129  &0.271   & 25.908  \\
\hline
Two w$\slash$ VC  & 3.248  & 0.232 & 15.229 \\
\hline
One w$\slash$ GC & 3.172 & 0.228 & 17.254 \\
\hline
Full w$\slash$o adv  & 2.404  & 0.415  &  81.980\\ 
\hline
Full w$\slash$ adv  & \textbf{3.419} & \textbf{0.216} &  \textbf{12.635}  \\
\hline
\end{tabular}
\end{center}
\end{table}

\begin{figure*}[!htp]
\centering
\includegraphics[width=0.95\linewidth]{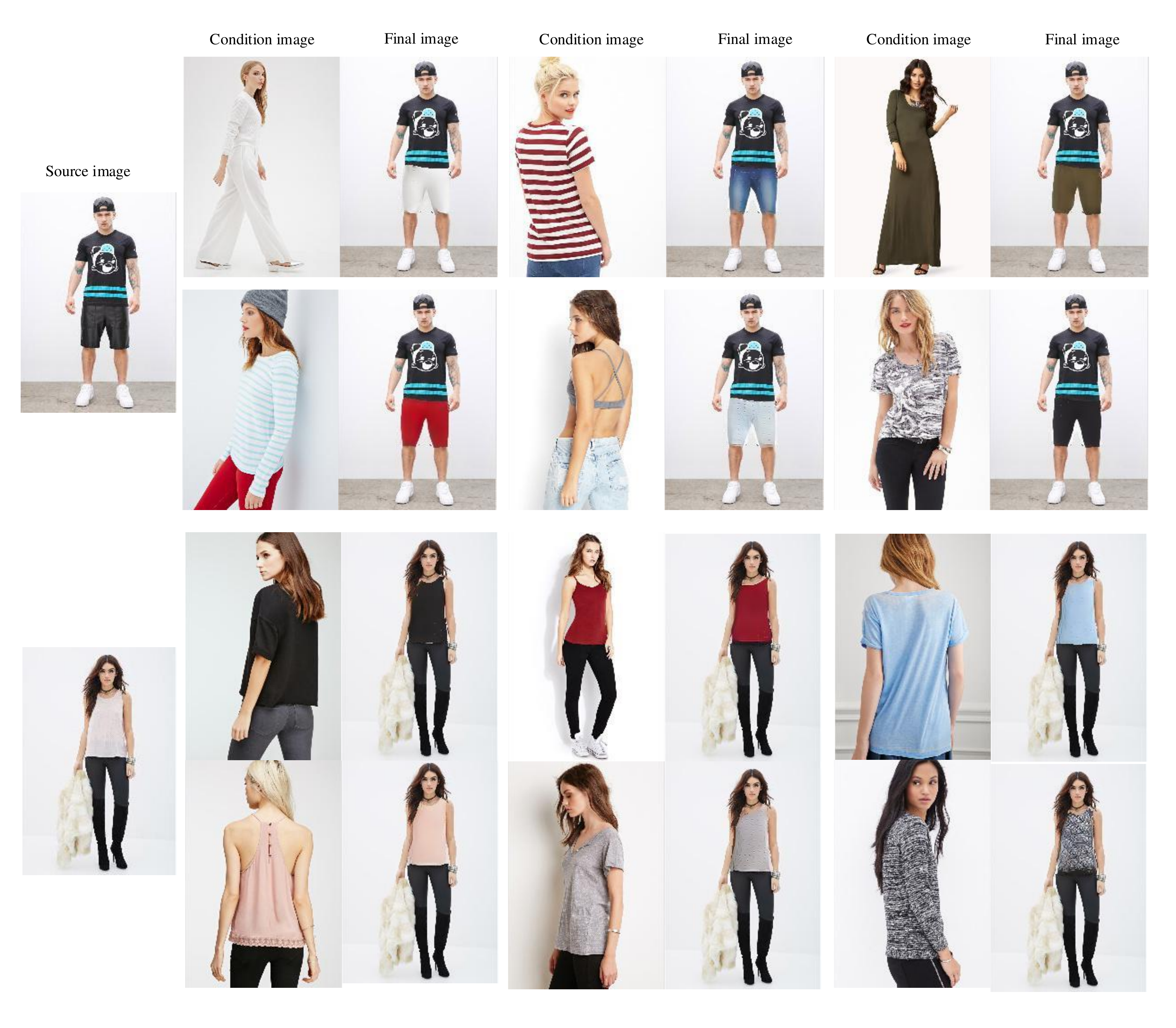}
\caption{Results of texture transfer using our method. From top to down, the first two rows show the results that transfer the texture of pants in condition images 
to the source images. The last two rows show the results that transfer the texture of upper clothes in condition images to the source images.}
\label{texture}
\end{figure*}

\subsection{Results analysis}
To give an intuitive demonstration on how the gated convolution works in parsing generator and image generator, 
we visualize the attention masks in all the gated convolution layers in Figure \ref{att}. 
The first column shows the input we need to deform, and the last two columns show the output and the target ground truth. 
The other columns show the attention masks in each gated convolution layer of the parsing generator and the image generator. 
The parsing generator aims to deform the source parsing, and the image generator is used to deform the source image with the guidance of parsing maps. 

As shown in Figure \ref{att}, feature selection, \ie, attention mask, varies from coarse to fine in both image and weight spaces among different layers. 
Specifically, feature selection focuses on background and large regions in the first layer while pays more attention to foreground and small regions in the small-scale layer.
Moreover, the weight variation between different semantic regions becomes smaller as the scale becomes smaller, but weight values vary in the opposite direction. 
This demonstrates that our multi-layer gated convolutions are able to dynamically select and deform the features from coarse to fine. 
Besides, the features aligned with the source parsing have large weights in the large-scale layer but have small weights in the small-scale layer. 
At the same time, the weights of features aligned with the target pose become lager as the scale becomes smaller. 
This demonstrates that feature selection and feature deformation are achieved together with different importance in different layers.

\subsection{Texture Transfer} 
As described in Section \ref{sec_trans}, our model can also achieve texture transfer. 
Figure \ref{texture} shows some visual results. The upper clothes and pants of the input person image can be automatically edited by using the texture of those 
in the condition images. Our model can edit the images by generating realistic components.

\subsection{Failure Cases}
\begin{figure}[!ht]
\centering
\includegraphics[width=1.0\linewidth]{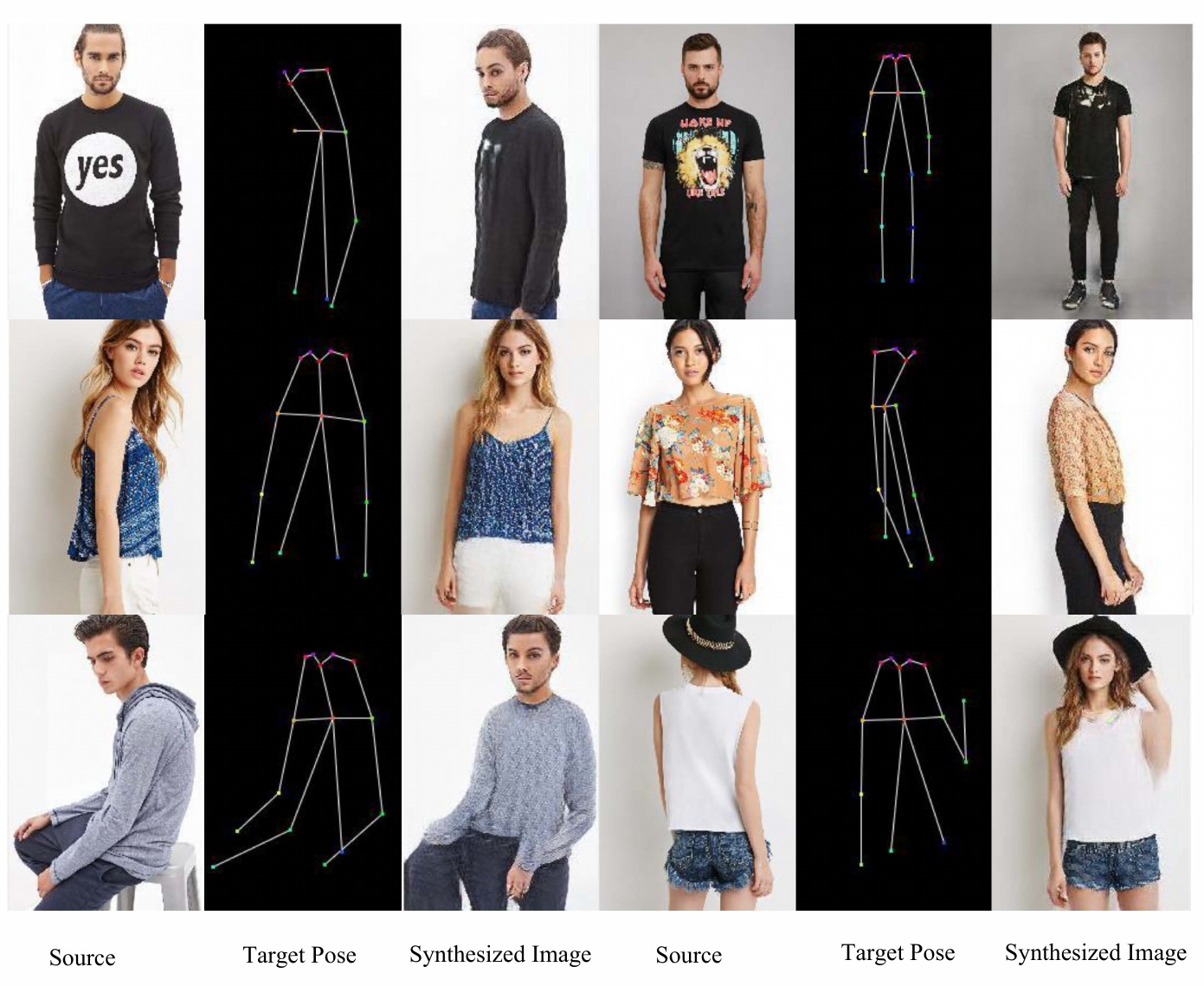}
\caption{Failure cases of our method.}
\label{fail}
\end{figure}

Our model can generate images that preserve the semantic shape and texture from the source image. 
However, as shown in Figure \ref{fail}, there are some limitations in our work. 
For the source images with complex texture (in the first and second rows), it is insufficient to use semantic correspondence extracted from human parsing maps as guidance. 
Besides, our training set does not have enough images including the specific carton patterns (in the first row), which is also the important reason. Moreover, for some complex poses, the synthesized images may be blurry.

\section{Conclusion}
\label{sec:conclude}
In this paper, instead of directly synthesizing the target image, we propose an adaptive hierarchical deformation framework for human pose transfer. 
The first deformation level generates human semantic parsing aligned with the target pose, 
and the second deformation level generates the final textured person image in the target pose with the semantic guidance.
Furthermore, we notice that the vanilla convolution is not suitable for unaligned generation task, 
and hence we use gated convolution to dynamically select important features and adaptively deform the image layer by layer. 
Experimental results demonstrate that our method achieves better pose transfer results with fewer parameters. 
Besides, our model can transfer clothing texture based on component attribute. 

\vspace{-3mm}
\section*{Acknowledgements}

This work was supported in part by Tianjin Research Program of Application Foundation and Advanced Technology (18JCYBJC19200).

\vspace{-.3cm}

\bibliographystyle{eg-alpha-doi} 
\bibliography{egbibsample}       



\end{document}